\documentclass{article}

\usepackage[preprint,nonatbib]{neurips_2020}

\usepackage[utf8]{inputenc} 
\usepackage[T1]{fontenc}    
\usepackage{hyperref}       
\usepackage{url}            
\usepackage{booktabs}       
\usepackage{amsfonts}       
\usepackage{nicefrac}       
\usepackage{microtype}      

\usepackage{graphicx}
\usepackage{caption}
\usepackage{subcaption}
\graphicspath{ {./figs/} }

\title{Concept-Modulated Model-Based Offline Reinforcement Learning for Rapid Generalization}

\author{
  Nicholas A. Ketz, Praveen K. Pilly\\
  Proficient Autonomy Center, Intelligent Systems Laboratory\\
  HRL Laboratories\\
  Malibu, CA 90265 \\
  \texttt{nick.ketz@gmail.com, pkpilly@hrl.com} \\
  }
\begin{document}

\maketitle

\begin{abstract}
The robustness of any machine learning solution is fundamentally bound by the data it was trained on.  One way to generalize beyond the original training is through human-informed augmentation of the original dataset; however, it is impossible to specify all possible failure cases that can occur during deployment. To address this limitation we combine model-based reinforcement learning and model-interpretability methods to propose a solution that self-generates simulated scenarios constrained by environmental concepts and dynamics learned in an unsupervised manner. In particular, an internal model of the agent's environment is conditioned on low-dimensional \emph{concept} representations of the input space that are sensitive to the agent's actions.  We demonstrate this method within a standard realistic driving simulator in a simple point-to-point navigation task, where we show dramatic improvements in one-shot generalization to different instances of specified failure cases as well as zero-shot generalization to similar variations compared to model-based and model-free approaches.
\end{abstract}

\section{Introduction}
\label{intro}
Many offline reinforcement learning (RL) solutions improve data efficiency by learning an unsupervised model of the agent's environment such that it can be used to generate samples to supplement the agent's true interactions with the environment.  These models can then be used in various ways to inform the agent's decisions, including offline training \cite{kaiser2019}, self-play \cite{silver2018}, and forward planning \cite{levine2020offline}.  Parallel to this field are the efforts to improve human understanding and interpretation of neural network outputs, often referred to as explainability (e.g., \cite{greydanus2018visualizing}).  

Our approach takes advantage of these two fields of study to improve RL performance and data efficiency by augmenting its own training through a learned concept-modulated model of the training environment.  The term World Model is analogous to a forward model in model-based RL literature \cite{moerland2020}, which provides an action-conditional prediction on the state-space of the agent.  The main focus within this work is to improve the World Model's predictive generalization within the environment.  We approach this problem by learning the World Model in an unsupervised fashion while also making it sensitive to the agent's goals and intentions though the use of \emph{concepts}.  These concepts are clusters over the agent's state-space that are empirically determined to be salient for the agent's action selection.  

The major contributions of this work are:
\begin{itemize}
    \item provide a method for unsupervised extraction of input-space clustering sensitive to an RL agent's action selection;
    \item improve World Model predictive accuracy by using these clusters as labels on the input-space, i.e., \emph{concepts}; and
    \item improve model-based RL agents' generalization though the use of simulated experiences based on the concept-modulated World Model.
\end{itemize}

\section{Related Work}
\label{rel_work}
\subsection{Offline Reinforcement Learning}
Much progress has been made on adapting RL methods to work in an offline fashion on a limited set of data.  Soft Actor-Critic \cite{haarnoja2018soft} and Conservative Q-Learning \cite{kumar2020conservative} are prime examples of this approach that are focused on reformulating methods that assume the training data is collected online by the learning agent to allow for training using data collected from other means (i.e., off-policy) while maintaining the efficiency of on-policy learning.  

Similarly, model-based RL methods try to learn a model of the target environment using self-supervised learning methods, which can be much more data efficient than model-free RL algorithms \cite{ha2018recurrent, hafner2020mastering}.  Once trained these models can be queried by the RL agent in an offline fashion to continue learning beyond the original data.  This process can continue iterating where more data is collected, the model is updated, and the agent learns from the updated model, which then yields more data for the next round of updates. 

\subsection{Data Augmentation}
Data augmentation methods, similar to those used in computer vision approaches, are being used for deep RL where images are a portion of the state-space \cite{kostrikov2020image,yarats2021mastering}.  Here the goal is to apply random shifts on the input image while enforcing that the learned Q function remain constant across them. This is similar to our approach, in that new data is generated to provided improved generalization performance; however, we use a learned model to generate that content rather than a limited set of prescribed image transformations. 

\section{Self-Generated Simulations}
\begin{figure}
    \centering
    \includegraphics[width=\textwidth]{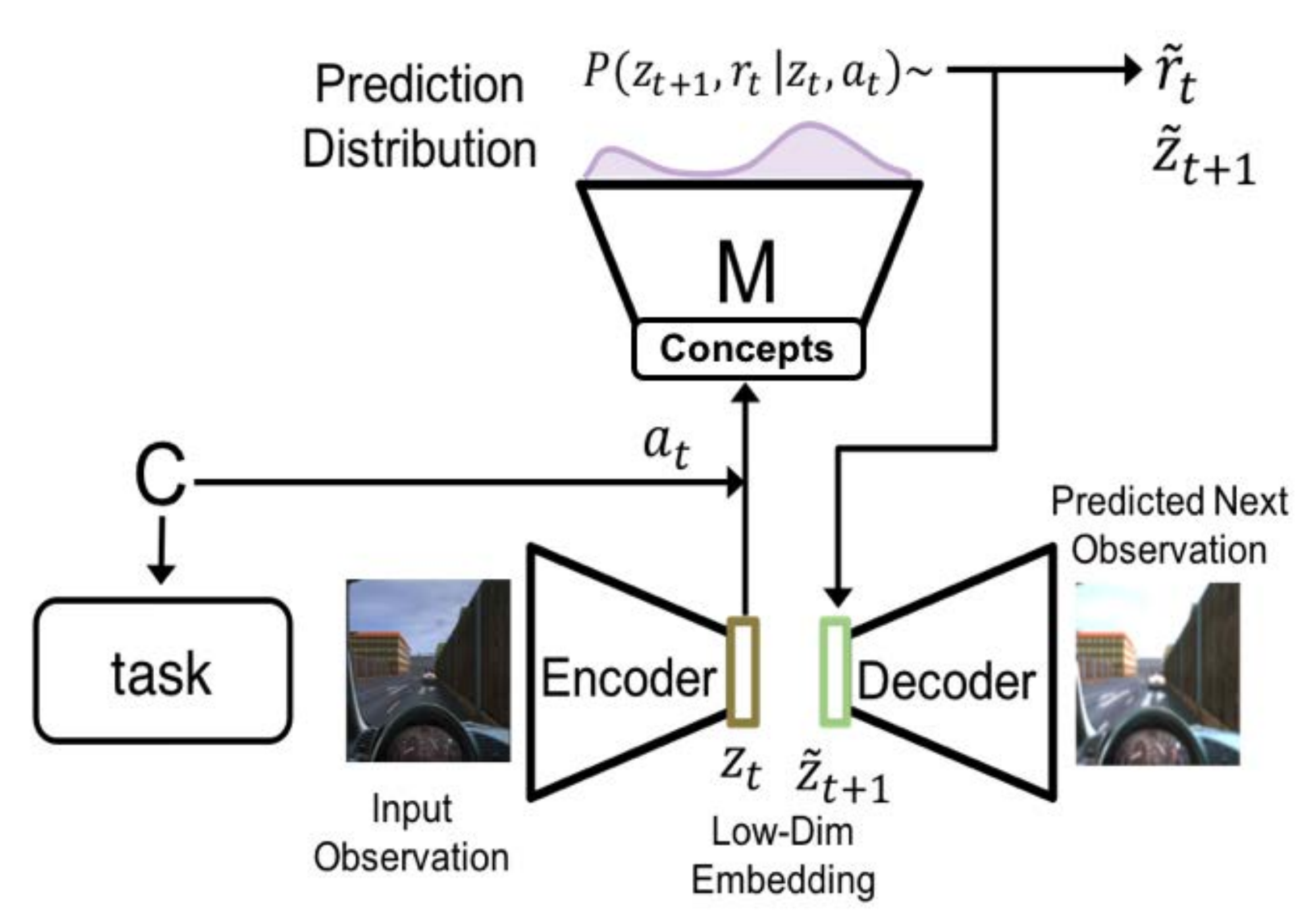}
    \caption{Block diagram of the full concept-modulated system, which is composed of an image-based input into a VAE that creates a low-dimensional embedding $z_t$, which is then combined with action $a_t$ from the Controller $C$ and the concept representation as the input into the Mixture Density Network World Model ($M$) to predict the distribution over potential next states and rewards ($z_{t+1}$, $r_{t}$).}
    \label{fig:sys}
\end{figure}

Here we describe the neural network supporting the offline training setup for our agent (Figure \ref{fig:sys}).  Our architecture is based on previous work in model-based RL \cite{ha2018recurrent} with the exception of a multi-layer perceptron policy optimized by gradient descent in the Proximal Policy Optimization (PPO) algorithm. Additionally, visual concepts are extracted using agent-determined saliency and leverages those concepts for augmenting the latent embedding of the state-space (Section \ref{concept}). We train a model of the environment composed of a Variational Auto-Encoder (VAE) as a convolutional image encoder/decoder and a Mixture Density Network (MDN) as a temporal prediction model.   These two components are trained separately on a dataset of episodes comprised of a sequence of state, action, reward, and episode done tuples from time 0 to $T$: $[s, a, r, d]_0^T$.  Once trained the components are used together to perform the interactive simulation of the agent's environment to support offline training. 

Briefly, the VAE is composed of a three-layer convolution network that encodes the input state (e.g., images) into a relatively low-dimensional latent representation ($z$), and a similar three-layer de-convolution network as decoder to reconstruct the original state.  This network is trained using the standard variational loss with a prior on the multi-dimensional latent distribution centered at 0 and with diagonal co-variances of 1, in addition to the reconstruction loss comparing the mean-squared error between the image input to the encoder and the image output from the decoder \cite{kingma2013}.  

The MDN takes the latent representation ($z_t$) at time $t$ as input to a Long Short-Term Memory (LSTM) layer.  The LSTM output is then concatenated with a one-hot encoding of the corresponding action ($a_t$) at that time step to create the input ($x_t$) to a fully connected linear perceptron layer, which outputs mean $(\mu_i)$ and standard deviation $(\sigma_i$) values used to determine a collection of Normal distributions ($N(\mu_i,\sigma_i)$), and a set of mixture parameters ($\Pi_i$) used to weight those separate distributions. This network is analogous to a Gaussian Mixture Model predicting the probability of the next state ($z_{t+1}$) provided the current state ($z_t$) and action ($a_t$): $P(z_{t+1} | z_t, a_t) = \sum_i(\Pi_i(x_t) N(\mu_i(x_t),\sigma_i(x_t)))$. The log likelihood of the parameters of this model can be fit, given some data, as a loss function ($L_{GMM}$) amenable to gradient descent \cite{bishop1994}.  Two additional output units are present in the MDN, one for predicted reward and the other for the predicted episode done state.  These are fit using a mean-squared error ($L_{MSE}$) and binary cross-entropy loss ($L_{BCE}$), respectively.  The network's parameters are optimized to minimize the summation of these three loss terms: $L = L_{GMM} + L_{MSE} + L_{BCE}$.

As noted above, in departure from \cite{ha2018recurrent}, a stochastic gradient descent based RL controller ($C$) was used for action selection. The $C$ network takes as input the current latent vector ($z$). Internally there is a single fully connected hidden layer that splits off to the Actor and Critic heads. Though we chose the PPO algorithm \cite{schulman2017proximal} for learning to illustrate the effectiveness of our approach in a policy gradient based method, any architecture in theory would work.  

\begin{figure}
    \centering
    \includegraphics[width=\textwidth]{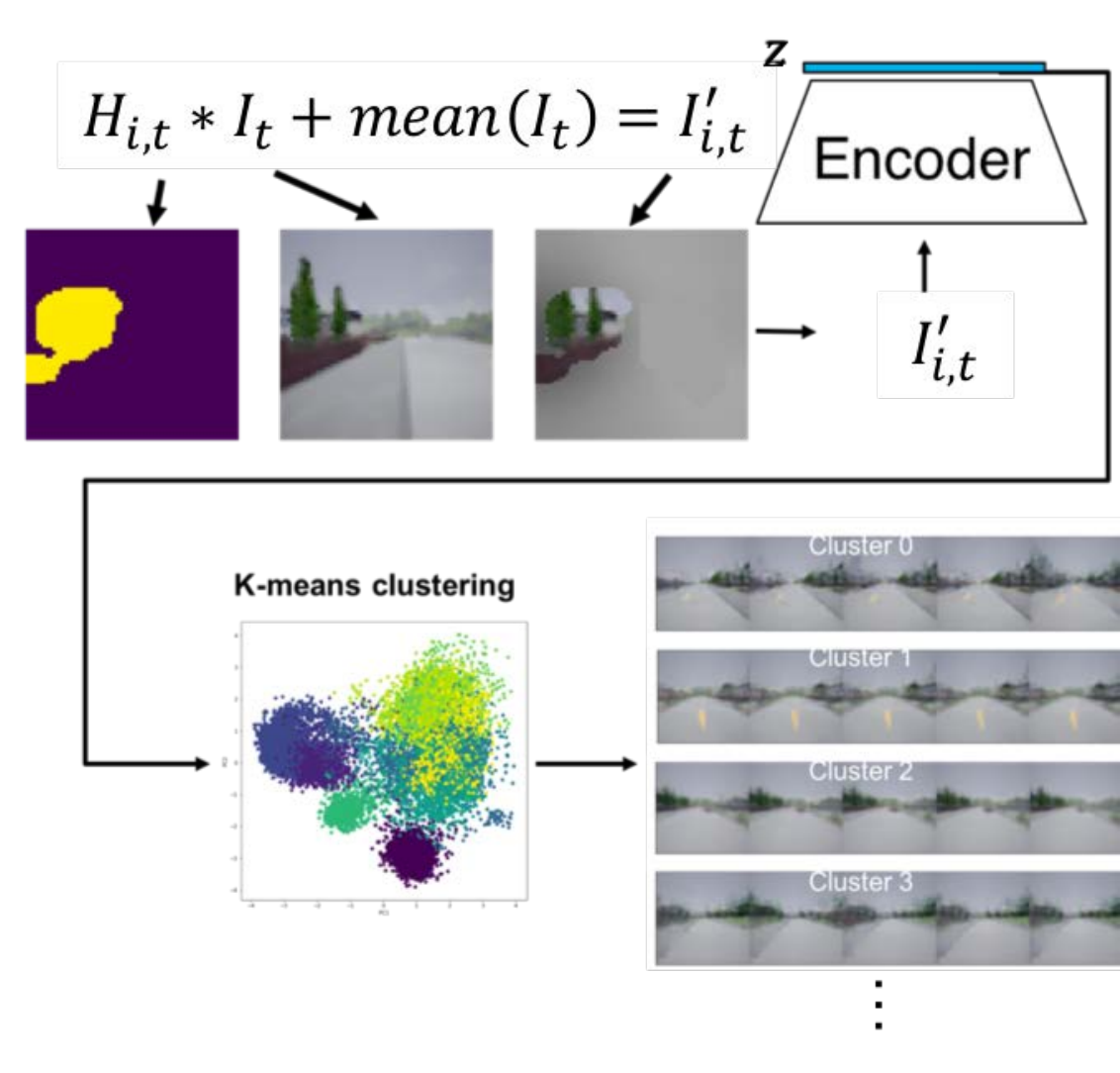}
    \caption{Concept extraction illustrates how agent-determined saliency is used to mask input images, encode them, and then cluster those encodings into visually similar clusters in the latent space of the encoder.  1 minus the normalized distances to these cluster centers are used as the concept representation when input to the $M$ network.}
    \label{fig:cm}
\end{figure}

\section{Visual Concept Extraction}
\label{concept}
In this section we describe the extraction of visual concepts from the agent's input stream, and their use in offline simulation generation. The term ``concept'' used throughout this paper generally means an additional low-dimensional representation of the input space. Our method for extracting this representation is to first determine portions of the input space that are relevant for an agent's action selection. We use perturbation methods similar to \cite{greydanus2018visualizing} to provide a saliency score for each pixel of each frame within a dataset of observations.  Saliency values are then z-scored and filtered for each observation such that values less than 2 standard deviations from the mean are set to 0. Connected components are extracted using the OpenCV library \cite{opencv_library}, and components that contain less than 1\% of the total pixels in the image are discarded.  This yields a dataset of $N*M_t$ salient visual patches where $N$ is the total number of observations and $M_t$ is the total number of connected components from the observation at time $t$. 

As shown in Figure \ref{fig:cm}, each connected component (indexed by $i$) within each observation is converted to a binary mask of the salient portion of the image ($H_{i,t}$) that is multiplied with the original image $I_t$, and finally the mean pixel value is added across that image. Each image in this dataset ($I^{\prime}_{i,t}$) has agent-determined salient portions of the image undisturbed at their original spatial locations situated on an uninformative background. This dataset of salient visual components is then input to the encoder network to get the latent representation for each component.  These encoded components, which now represent points in the encoder's latent space, are then clustered into $N$ clusters using the K-means algorithm.  The specific number of clusters used here was 10 based on a loose fit using the elbow method, though its impact on performance has not been well explored.  This clustering effectively segments the encoder's latent space according to the agent's visual policy, providing potentially high-level descriptors of image patches that are relevant to agent behavior. This is visualized on the lower right side of Figure \ref{fig:cm}, where five samples from four of the 10 clusters are shown. Here the clusters can be seen with some interpretable consistency; e.g., Cluster 0 seems to represent normal driving, Cluster 1 when the agent crosses the center line, Cluster 2 an oncoming obstacle, and Cluster 3 when the agent is encroaching on the left lane. 

We then train an $M$ network based on its standard input concatenated with the cluster labels.  Specifically, the input is the concatenation of the latent representation $z_{t}$ of the current observation, a one-hot encoding of the current action $a_{t}$, and 1 minus the normalized distance of the current encoded observation to each cluster center. Here the normalization imposes that the cluster furthest from the current observation will have a value of 0, and the value of the closest cluster will be 1.

\section{Experiments}
\label{exp}
Experiments are implemented within the CARLA autonomous driving simulator using the point-to-point navigation task within Town01 \cite{dosovitskiy2017carla}. Each time step in the environment provides an RGB camera image from the driver's perspective. The reward function for the navigation task is a simplified estimate of safe and effective driving from some starting point to a target end point:

$r_t=1000(d_{(t-1)}-d_t )+0.05(v_t-v_{(t-1)} )-0.00002(c_t-c_{(t-1)} )-2(s_t-s_{(t-1)} )-2(o_t-o_{(t-1)})$

Reward ($r_t$) for a given time point $t$ is a weighted sum of five terms based on the distance to the goal $d_t$ in km, speed $v_t$ in km/h, collision damage $c_t$, intersection with the sidewalk $s_t$ (between 0 and 1), and intersection with the opposite lane $o_t$ (between 0 and 1). The python-based Multi Agent Connected Autonomous Driving (MACAD) gym \cite{palanisamy2020multi}, which interacts with the CARLA environment, was modified for these experiments. There are five starting and end point defined driving routes during training, and for a given instantiation in the environment with a specific starting and end point pair, the agent is allowed to interact with the environment for 300 time steps or until it reaches the prescribed end point. The time series data extracted from the agent’s interactions with the environment is saved as a list of $[s,a,r,d]$ values for each time step. This collection of time series data is throughout this paper referred to as a ``rollout'' of experience or an ``episode”.

An initial dataset is first collected from the CARLA simulator through the interactions of a randomly initialized deep RL neural network with the same architecture as the original DQN \cite{mnih2015human} and optimized using the PPO algorithm for 130,000 environment interactions over the five unique starting points.  These saved interactions, each comprised of a $[s,a,r,d]$ tuple, yielding several hundred episodes each with a maximum length of 300 interactions. 

Using this dataset for offline training of VAE and MDN, we have currently investigated the influence of concept-modulation in two experimental settings: unspecified generalization where we measure generic performance improvements following limited experiences, and specified generalization where we measure performance in executing a particular scenario given limited exposure to that scenario in addition to the original limited experiences.

\subsection{Unspecified Generalization}
\label{unspec}
This section outlines experiments done to improve performance generally given limited experiences. Sample efficiency is defined as the level of performance provided a fixed budget of environment interactions and used as the target metric to assess the effectiveness of our approach.  Here we use 100,000 interactions as the budget of online interactions. The full system is trained in sequence: VAE, then MDN, then Controller. We compare performance on three types of trained agents: Model-Free (MF), Model-Based (MB), and Concept-Modulated (CM). 

VAE training was done using a network with a latent dimension size of 64 units and a batch size of 32 samples for up to 500 epochs of the full 100,000 samples.  Early stopping of training was used such that training stopped when the test error (evaluated on the last 30,000 samples collected that were not used during training) showed no decrease in test loss greater than $10^{-4}$ for 30 contiguous epochs. The trained VAE was then used as an encoder on the states to be fed into the MDN. Each epoch of MDN training was done using full episodes in batches of 20 over the whole 100,000 frames for up to 500 epochs.  Early stopping was again triggered if test error did not decrease by $10^{-4}$ for 10 contiguous epochs.  

Finally, the PPO Controller training was done using a minimal architecture that consisted of a fully connected network with a hidden dimension of 512 units that split into two output heads: the nine discrete actions for the actor head and the single value output for the critic head.  As noted above, three separate agents were trained using this architecture.  The first agent, referred to as model-free (MF), is most similar to standard deep RL and trained online based on live interactions with the CARLA simulator for 100,000 steps while using the pre-trained VAE as an image encoder.  The model-based (MB) and concept-modulated (CM) agents continued training from where the MF agent stopped.  By using the pre-trained VAE and MDN to create offline simulations of the environment that don't count against the training budget, MB and CM agents continued to optimize the weights in the C network. The CM agent used samples generated from the VAE and a MDN that was trained with the concept representation concatenated onto its inputs as described above, while the MB agent used no concepts.  

Offline training of MB and CM agents was based on 1 million interactions with the standard and concept-modulated World Models, respectively.  These actor-critic models used the parameters of the MF agent as the starting point, and their ratio clipping hyperparameter for PPO was reduced by 10x to 0.01 to keep gradient updates relatively small. Seeding of the MDN is done using a sequence of 10 contiguous frames selected from a random starting point within a randomly selected episode from the stored dataset of interactions for the MF agent. Here the encoded $z$ vector for a given frame is provided with the original action $a$ as input to the MDN for the sequence of seed frames, after which the outputs of the MDN and MF agents are used to continue the simulation of the environment until the MDN output for done state $d$ is greater than 0.5.

\subsection{Specified Generalization}
This section outlines experiments done to test performance on a targeted failure case where a single episode of that failure case is provided as additional training data.  Here the particular failure case was designed such that a leading vehicle that is stationary for approximately 5 s is created at the start of an episode to be within a given distance to the ego vehicle.  Failures due to collisions with the obstacle vehicle occur in this case when the target distance is sufficiently close to the ego vehicle such that the original dataset has no examples of these scenarios.  This failure case is designed to illustrate the relatively brittle nature of the learned policies from model-free RL and the comparatively more robust nature of model-based methods. It also mimics the most dominant scenario leading to car crashes as documented by the National Highway Traffic Safety Administration (NHTSA) \cite{wassim2007}.

To investigate specified generalization, we used the same learned parameters for the VAE, MDN, and the three agent types described in Section \ref{unspec} as a starting point. A single episode of a leading obstacle vehicle created 10 m from the ego vehicle was logged for each of the five starting points. These five episodes were then exclusively used as seed data to generate content for offline training as described above.  

\section{Results}
\label{results}
\begin{figure}
     \centering
     \begin{subfigure}[b]{0.45\textwidth}
         \centering
         \includegraphics[width=\textwidth]{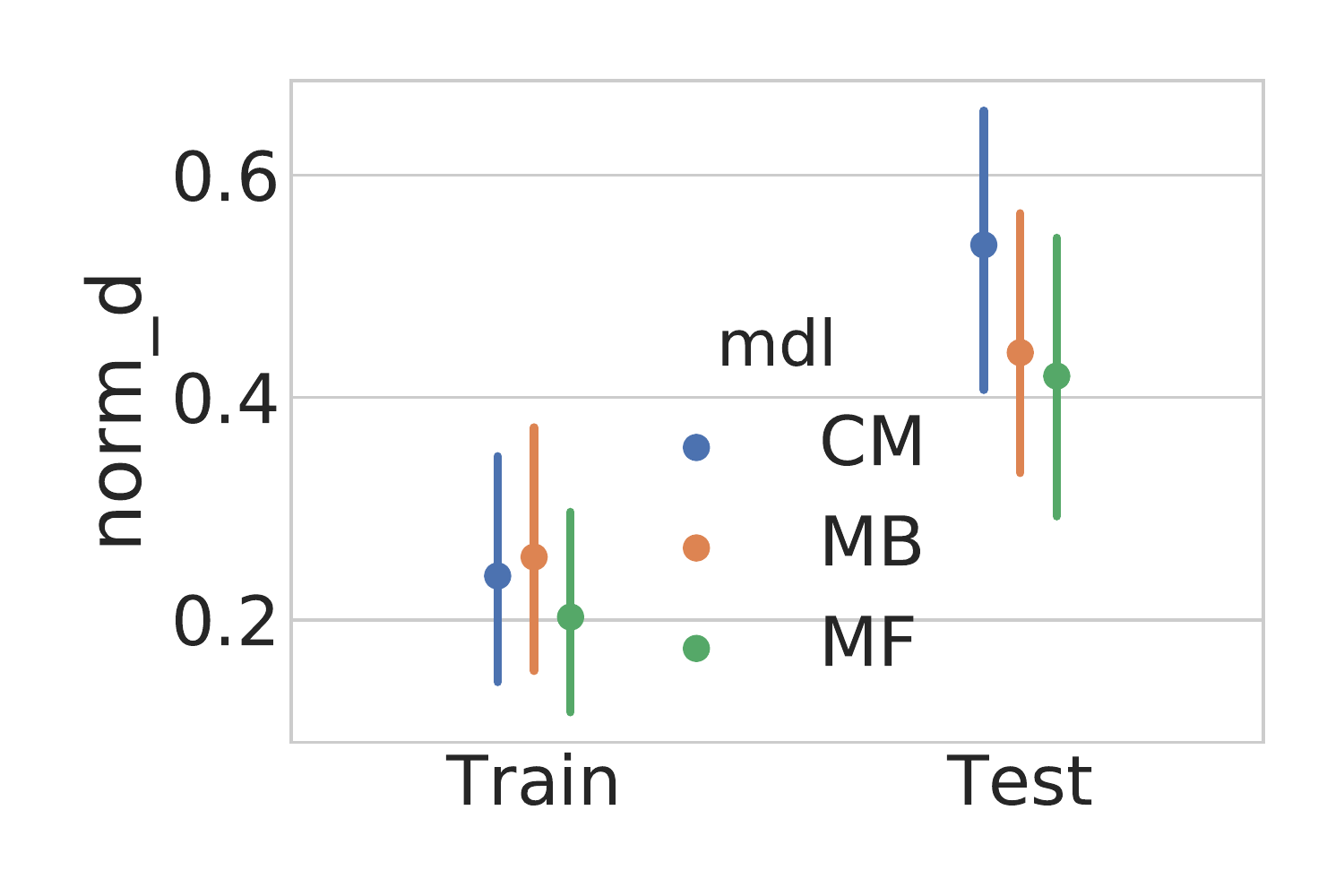}
         \caption{Normalized Distance to Goal}
         \label{fig:norm-d}
     \end{subfigure}
     \hfill
     \begin{subfigure}[b]{0.45\textwidth}
         \centering
         \includegraphics[width=\textwidth]{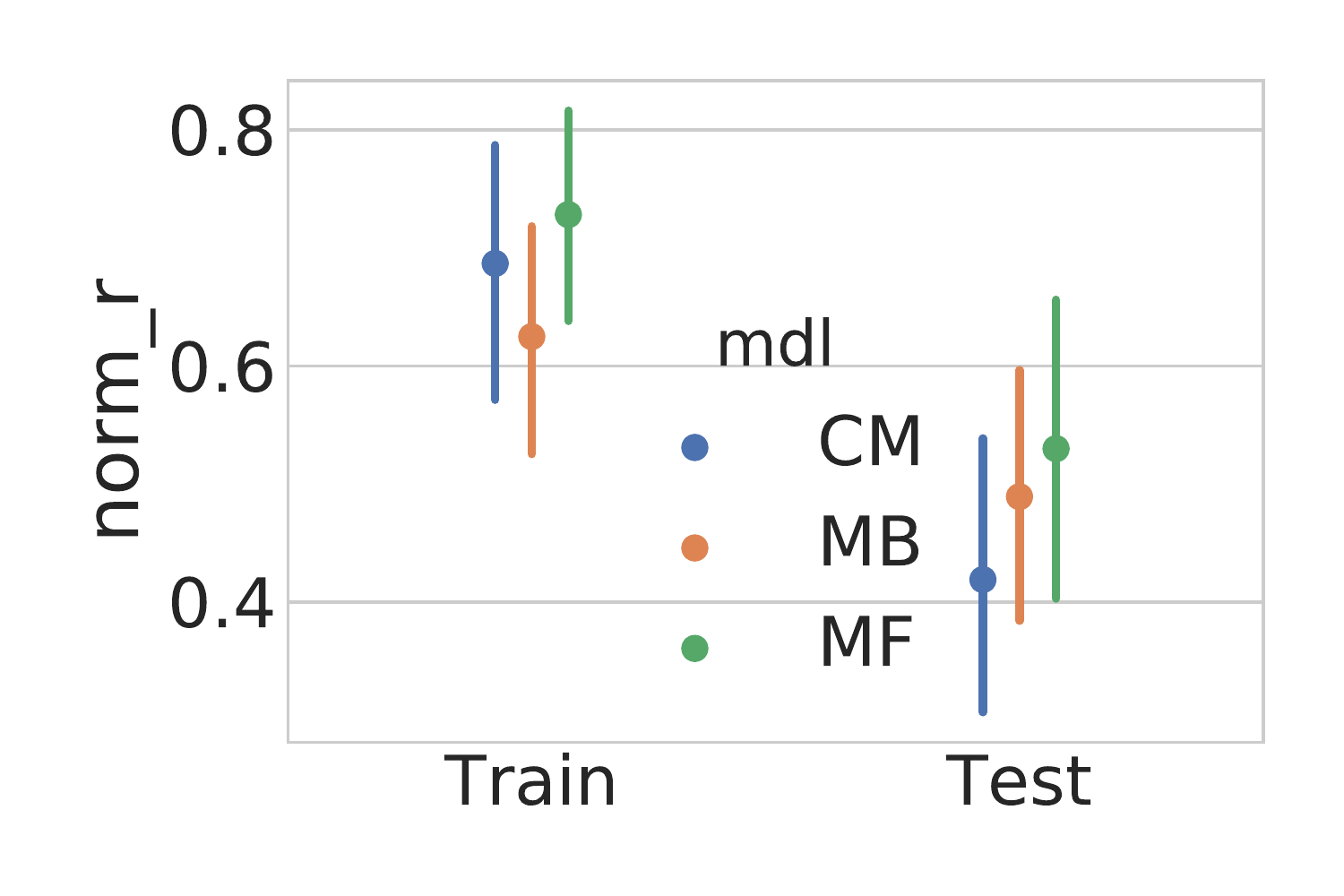}
         \caption{Normalized Cumulative Episode Reward}
         \label{fig:norm-r}
     \end{subfigure}
     \caption{Performance on unspecified generalization for two metrics, normalized distance to goal (norm\_d) and normalized cumulative episode reward (norm\_r), for the three agent types: Model-Free (MF), Model-Based (MB), and Concept-Modulated (CM). Performance data points marked as Train are from the five starting points that were included in training, while those marked as Test are derived from five other starting points that were not included.  Error bars are 95\% bootstrap confidence intervals.  No clear differences among agent types is seen.}
     \label{fig:perf_ug}
\end{figure}

\subsection{Unspecified Generalization}
\label{unspec-result}
Here we show results related to generic performance improvements for MB and CM agents compared to the MF agent. As seen in Figure \ref{fig:perf_ug}, performance is plotted as the normalized distance to goal at the end of episode (a distance of 1 implies no change from the start and distance of 0 implies the goal location), as well as the normalized cumulative episode reward (0 implies no reward and 1 the approximate maximum episode reward).  Mean performance values for the three agent types are plotted; namely, MF, MB, and CM.  Points labeled as Train come from the five starting points that were included in training, while points labeled as Test come from five other starting points that were not included. 

Both distance and reward metrics show a degradation of performance from training starting points to held-out test starting points; however, there is no significant difference among the agent types within either train or test starting points for both metrics.  The general conclusion drawn from these results is that offline training does not seem to provide data efficiency benefits beyond the model-free approach for unspecified generalization.  

\subsection{Specified Generalization}
Here we show results related to improvements in one-shot generalization to the particular failure case for MB and CM agents compared to the MF agent.  This specified failure case, as noted in Section \ref{exp}, is a collision with an obstacle vehicle that appears 10 m in front of the ego vehicle. Success in this scenario is defined as avoiding collision with the obstacle vehicle.  In particular, we measure the probability of successfully avoiding collision with the obstacle vehicle across five episodes from the five trained starting points and three training runs, yielding a total of 75 episodes per agent type as the normalizing constant. 

\begin{figure}
     \centering
     \begin{subfigure}[b]{0.45\textwidth}
         \centering
         \includegraphics[width=\textwidth]{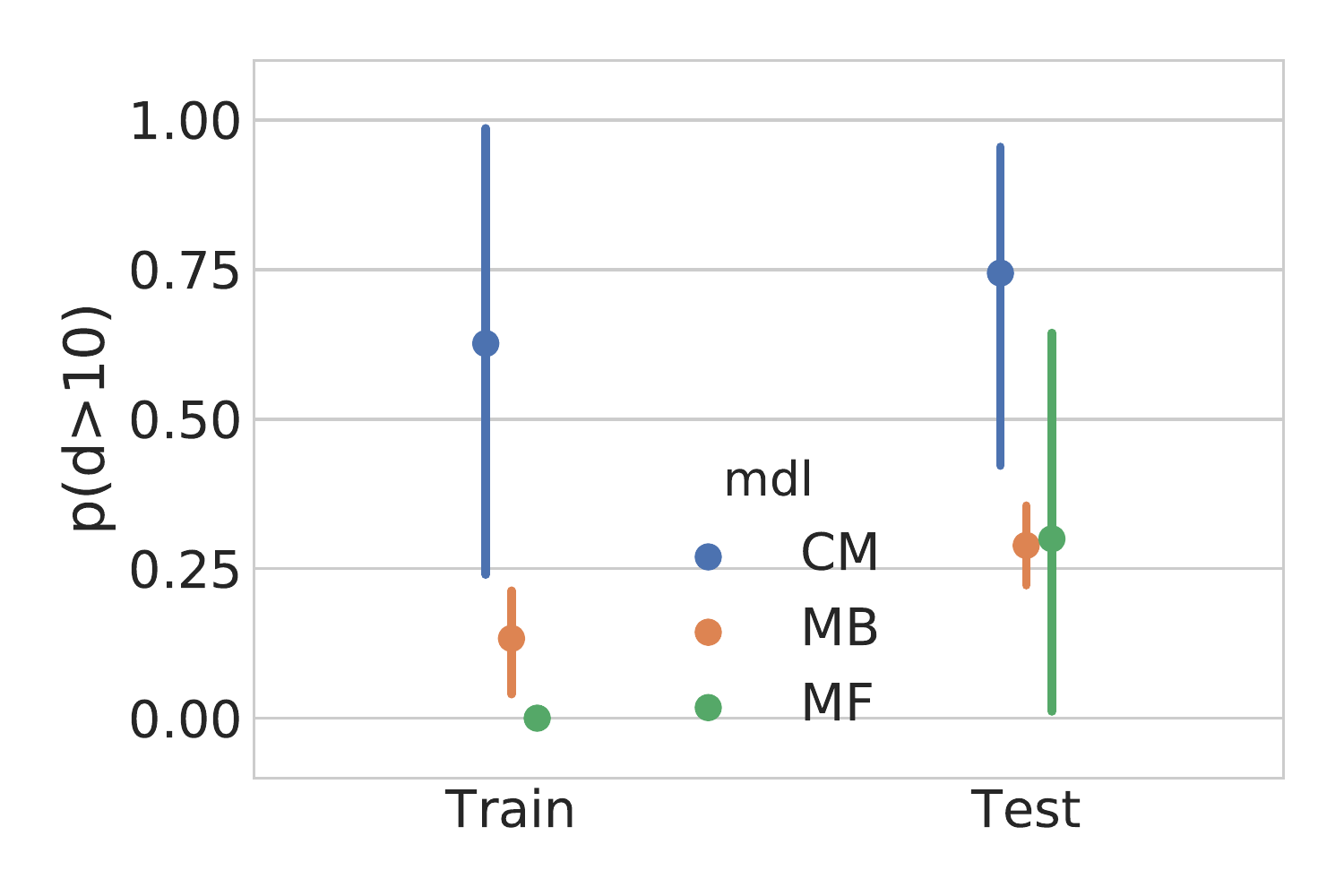}
         \caption{Obstacle at 10 m}
         \label{fig:10m}
     \end{subfigure}
     \hfill
     \begin{subfigure}[b]{0.45\textwidth}
         \centering
         \includegraphics[width=\textwidth]{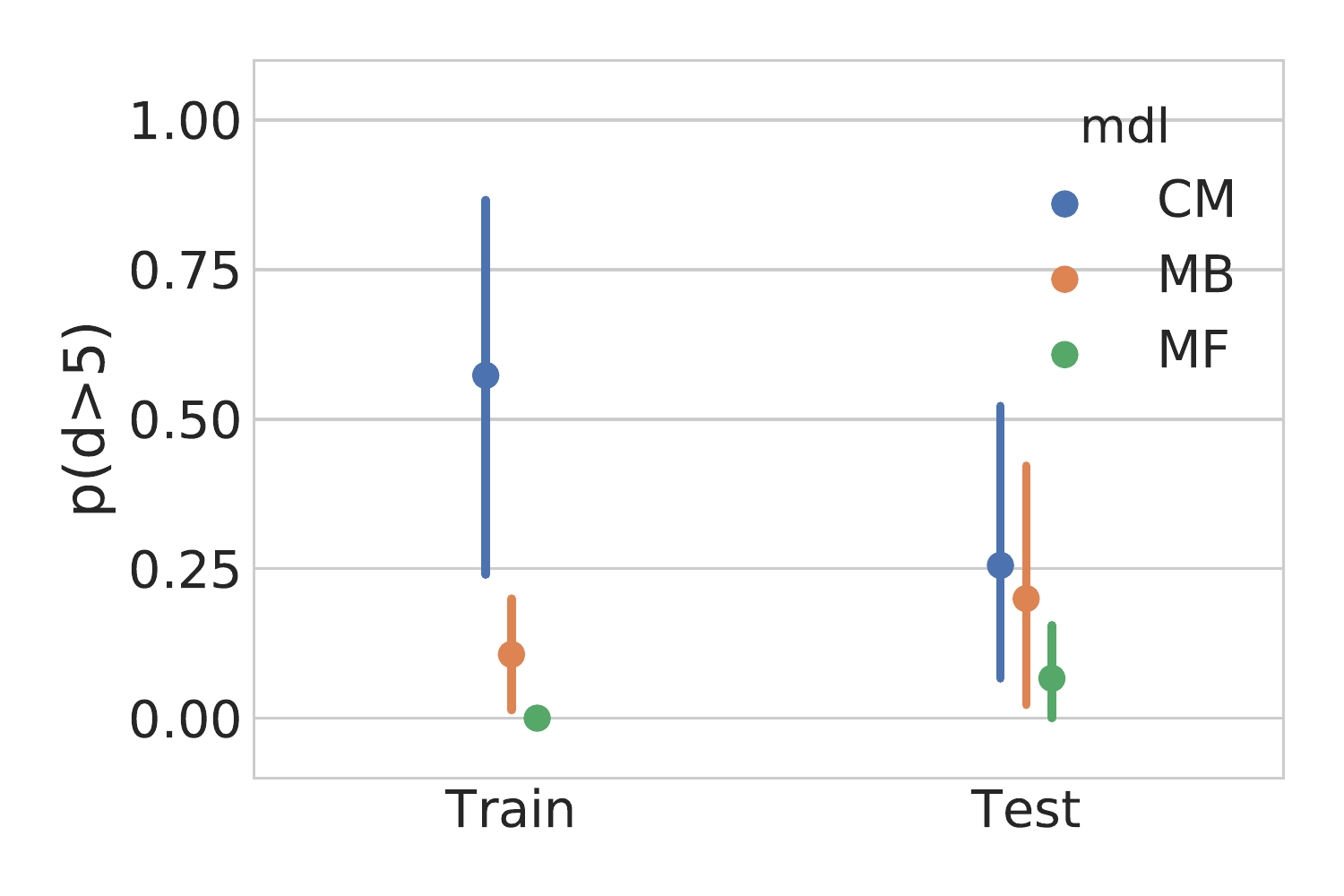}
         \caption{Obstacle at 5 m}
         \label{fig:5m}
     \end{subfigure}
        \caption{Generalization performance for two obstacle distances, 10 m and 5 m.  Probability of successfully avoiding collision with the obstacle is plotted on the y-axis for the three agent types: MF, MB, and CM, where MB and CM agents are trained offline based on a single episode with the obstacle at 10 m.  Points labeled as Train are derived from the ego vehicle’s five starting points that had a failure case example episode to seed simulated rollouts, while those labeled Test did not. Error bars are 95\% bootstrap confidence intervals.}
        \label{fig:gen}
\end{figure}

Figure \ref{fig:gen} shows the probability of successfully avoiding the collision for the three agent types and two obstacle distances.  Single episodes from the five training starting points were derived when the obstacle vehicle appeared 10 m in front of the ego vehicle, and evaluations following the offline training for this case are shown in Figure \ref{fig:10m}.  Here we can see the CM agent performs the best, both on the five starting points that were included in training (labeled Train) and on the five starting points that were never seen (labeled Test).  These results show the CM agent can do both one-shot generalization shown in the 10 m Train starting points, and even zero-shot generalization as illustrated in the 10 m Test starting points.

We also tested these same trained agents when the obstacle vehicle appeared at 5 m in front of the ego vehicle, which was present in neither the failure case data nor the training data. Performance in this scenario is analogous to zero-shot learning. As seen in Figure \ref{fig:5m}, the CM agent shows similar performance on the starting points that had an example obstacle vehicle at 10 m when evaluated on with an obstacle at 5 m.  In contrast, there is no clear difference among the model types for the starting points that had no example failure cases at 10 m when evaluated at 5 m. Here we can see evidence to suggest that concept-modulation is able to perform zero-shot generalization within some range of previous experiences and clearly exceeds the other two agent types. 

\section{Discussion}
This paper illustrates the principles of a concept-modulation mechanism within model-based RL to promote generalization to different instances and variations of specified failure cases.  This modulation mechanism is meant to extend model-based and model-free RL approaches to improve performance in low-data regimes. However, this improvement was not noticed for unspecified generalization in the case where 100,000 samples are used for model-free learning. Even the model-based approach showed little to no improvement in data efficiency over the model-free approach. There are a few potential reasons for this discrepancy with existing visual model-based RL approaches \cite{kaiser2019}. One would be the complexity and partial observability of the autonomous driving environment compared with the relatively easier Atari and MuJoCo environments. The other would be the potentially unstable nature of off-policy learning from model-based simulations of the environment.  We suspect this can be addressed by using better off-policy learning methods such as Soft Actor-Critic \cite{haarnoja2018soft}, but this is left for future work. 

Given our approach performs much better in generalization to the specified failure case, we posit that concept-modulation allows for more useful offline simulations of the single failure case in comparison to the model-based agent, which has no concepts. Our method can be used with any existing model-based approach provided the ability to project the agent-determined saliency on the input space. Presumably this can provide one-shot generalization improvements for existing state-of-the-art model-based and offline RL approaches such as \cite{yarats2021mastering, ye2021mastering, hafner2020mastering}; however, this remains to be tested. Future work should also evaluate concepts as a bottom-up clustering of the $z$ space rather than from the encoding of the agent-determined saliency patches.  This would provide insight and justification for the benefit of leveraging the agent-based saliency compared to the relatively simple alternative of clustering based on unfiltered input images. Finally, our concept-modulated generative modeling should be integrated with compatible continual learning approaches (e.g., \cite{ketz2019}) to additionally avoid catastrophic forgetting of previously acquired capabilities. Theoretically there should be a trade-off where the right balance of replays and generated samples promotes both generalization and preservation.

Work is also needed to validate and further boost the semantic grounding of content generated by the World Model with and without concept-modulation. We hypothesize that extracting visual concepts that are sensitive to agent's action selection and consolidating them for temporal predictions makes the generated sequences of frames more realistic in terms of visual and temporal coherence, which can be assessed using the human-validated Fr\'{e}chet Video Distance metric \cite{unterthiner2018}. 

\bibliographystyle{plain}

\end{document}